# Development of A Stochastic Traffic Environment with Generative Time-Series Models for Improving Generalization Capabilities of Autonomous Driving Agents

Anil Ozturk[1*], Mustafa Burak Gunel[2*], Melih Dal[3], Ugur Yavas[4] and Nazim Kemal Ure[5]

*Abstract*— Automated lane changing is a critical feature for advanced autonomous driving systems. In recent years, reinforcement learning (RL) algorithms trained on traffic simulators yielded successful results in computing lane changing policies that strike a balance between safety, agility and compensating for traffic uncertainty. However, many RL algorithms exhibit simulator bias and policies trained on simple simulators do not generalize well to realistic traffic scenarios. In this work, we develop a data driven traffic simulator by training a generative adverserial network (GAN) on real life trajectory data. The simulator generates randomized trajectories that resembles real life traffic interactions between vehicles, which enables training the RL agent on much richer and realistic scenarios. We demonstrate through simulations that RL agents that are trained on GAN-based traffic simulator has stronger generalization capabilities compared to RL agents trained on simple rule-driven simulators.

## I. INTRODUCTION

Autonomous driving systems improved significantly in the recent years due to the increasing computational capabilities and development of novel machine learning algorithms. Although fundamental maneuvers such as cruise control or lane keeping are almost mature technologies, automating more advanced maneuvers such as lane changing is still an open problem.

Automated lane changing from the perspective of operational decision-making has been studied by [1], [2], [3]. Such methods usually neglect the highway traffic conditions and treat the problem from a local perspective. In order to apprehend the complex dynamics of highway traffic, such as avoiding long term traffic jams, a strategic decision-making approach is necessary.

### A. Previous Work

Machine learning methods and rule-based methods are the two main methods that have been used in autonomous lane change problems. In recent years machine learning methods became more prominent due to their generalization capability and adaptation to real-world data. One of the more popular machine learning approaches in this context is reinforcement learning (RL) [4], where autonomous driving agents are trained on the traffic simulators to learn good lane changing policies. The agent in [5] is trained for producing lane changing and acceleration/deceleration actions using a deep RL approach. Compared with the rule-based approach, the agent shows a promising performance. However, the simulator used in the work assumes very simple maneuvers for surrounding vehicles, which does not fully reflect the complexity of real world scenarios.

In [6], automation of the lane change and speed adjustments have been achieved by a combination of deep reinforcement learning and Monte Carlo tree search algorithms. A neural network that utilizes convolutional and fully connected layers are developed for two different agents and compared against rule-based MOBIL [7] and Intelligent Driver Model (IDM) [8] methods. Authors' previous work [9] also utilizes a similar approach and shows that deep reinforcement learning approaches can outperform the rule-based approaches significantly in the presence of sensor and process noise in the environment.

One of the most fundamental gaps in the existing work is, RL agents exhibit significant simulator bias when they are trained on simple traffic simulators. Most existing work assume that surrounding vehicles employ rule-based decision making algorithms such as MOBIL and IDM. Hence the traffic surrounding the ego vehicle always follow smooth and meaningful trajectories, which does not reflect the real world traffic where surrounding vehicles driven by humans mostly execute erroneous maneuvers, hesitate during lane changes and overall perform much inferior compared to algorithms like MOBIL and IDM. Thus RL agents trained on such simulators usually do not generalize well to real world scenarios, because they are incapable of predicting erroneous maneuvers executed by human drivers.

### B. Contribution

Our main contribution in this work is the development of a data-driven traffic simulator, where we simulate trajectories of the surrounding traffic by using a generative adversarial deep neural network (GAN) trained on real traffic data. We show that generated randomized trajectories resemble real life scenarios and thus the developed simulator provides a much richer and realistic environment for training RL agents.

* These authors contributed equally to this work
[1]A. Ozturk is with Faculty of Computer Engineering, Istanbul Technical University, Turkey `ozturka18 at itu.edu.tr`
[2]M.B. Gunel is with Faculty of Aeronautics and Astronautics, Aerospace Engineering, Istanbul Technical University, Turkey `mustafa.gunel at itu.edu.tr`
[3]M. Dal is with Faculty of Computer Engineering, Bogazici University, Turkey `melih.dal at boun.edu.tr`
[4]U. Yavas is with Eatron Technologies, Istanbul, Turkey `ugur.yavas at eatron.com`
[5]N.K. Ure is with ITU Artificial Intelligence and Data Science Research Center and Department of Aeronautical Engineering, Istanbul Technical University, Turkey `ure at itu.edu.tr`

Next, we develop an RL agent for automated lane changing that is suitable for training on both GAN based and rule-based simulators. Our results show that RL agents trained on GAN-based traffic have significantly better generalization capabilities compared to agents trained on rule-based traffic simulators.

## II. BACKGROUND

### A. Data Driven Traffic Modelling in Highway Environment

Modelling traffic in a highway is a multi faceted problem. A single vehicle's motion can be thought of as a time series motion modelling problem, but other vehicles entering the scene creates an interactive environment where vehicles maneuvers affect each other. Therefore modelling efforts usually have two parts, one accounting for single agent motion modelling, the second one for context-awareness of the vehicle.

Motion modelling in a dynamic scene almost always has to include the interactions of other agents. For that purpose, an occupancy grid around the vehicle/object of interest is drawn and that grid is processed by different methods to come to a conclusion about the interaction with environment. Some of the popular approaches include graph based [10] or convolution based [11] models to extract interaction models. In this work, a time series interaction model is required and the network first suggested in [12] is used. In this method, a social pooling is introduced where a long short term memory (LSTM) Encoder encodes all the vehicles' position in a relative manner to the rest of the vehicles then a max pooling operation is performed at the hidden states of the encoder; arriving at a socially aware module.

### B. Reinforcement Learning

We formulate the lane change decision making problem as a Markov Decision Process and use Q-learning to compute policies that yield lane changing decisions while optimizing safety and performance. In the remainder of this section we provide a brief overview of MDPs and Q-Learning. Traditional reinforcement learning process is mostly based on Markov Decision Processes (MDP).

A MDP is defined by the tuple $(S, A, P, R, \gamma)$, where $\gamma$ is the discount factor, $R$ is the reward function, $S$ is set of Markov states, $A$ is the action space of the agent and $P$ is the state transition probability matrix. The discount factor $\gamma$ is set between $0 < \gamma \leq 1$. The next state of the agent is governed by the probabilistic transition determines by the current state and current action (see Eq. 1) The main purpose of the RL agent is to get maximum total reward in a long term (Eq. 2) by interacting with the environment by choosing an optimal policy $\pi : S \rightarrow A$.

$$P(s_{t+1} = s'|s_t = s, a_t = a) \quad (1)$$

$$E\left[\sum_{t=0}^{\infty} \gamma^t R(s_t, s_{t+1})\right] \quad (2)$$

The main objective of Q-Learning is to compute the value function $Q(s, a)$, which determines the long term total reward of taking action $a$ at state $s$. Hence if the optimal value function is determined, optimal policy $\pi$ can be obtained by taking the action $a = \arg\max_{a'} Q(s, a)$ at state $s$. Classic Q-learning algorithm iteratively updates the the value function estimate by applying the update in Eq. 3, where $\alpha_k$ is the learning rate.

$$Q(s,a) \leftarrow (1-\alpha_k)Q(s,a) + \alpha_k(R(s) + \gamma \sum_{s'} \max_{a'} Q(s', a')) \quad (3)$$

Classic Q-learning is not applicable to problems where storing all state-action pairs is $(s, a)$ is infeasible. Deep Q-Network (DQN) [13] algorithm approximates the $Q(s, a)$ by using deep neural networks, hence enabling computation of the value function for large-scale and continuous problems.

In this work, the *Rainbow-DQN* [14] method has been utilized, which combines several properties of the recent best improvements in DQN; *double Q-Learning*, *prioritized experience replay*, and *dual network architecture*. As the method requires, each experience (state,action transitions) and their outcomes are stored in to a buffer, and the value approximating Q-network is trained by randomly choosing from this set of experiences/outcomes and updating the network. Experiences in this buffer may vary in their importance; as more recent experiences and their outcomes are more relevant. The gradients learned from these experiences are multiplied by their *importance weight* to compensate for this mismatch, a concept from Monte Carlo Sampling. Interested readers are referred to original publication [14] for the details of the algorithm.

## III. TIME SERIES GENERATIVE ADVERSARIAL NETWORKS

Generative adversarial networks [15] have been widely used on image generation problems [16]. The idea is to develop a min-max game between two deep neural networks, where the **generator** tries to generate fake samples from some real data, and the **discriminator** tries to discriminate between real and fake samples. In the case of vehicle trajectories, generator takes the real vehicle trajectories and generate new trajectories, the discriminator classifies the generated trajectories as real or fake.

There exists different ways of adapting GAN architecture to time series data. One method is to convert the time series data into a 2D array and then perform convolution on the data as the original architecture suggests. In [17] time series data have been put into 2D format and the architecture in DCGAN [18] has been applied to the data. Another alternative, which is also used for this work, is to develop a sequence to sequence [19], encoder and decoder LSTM network. Such approaches have been extensively used in similar problems. In [20], the same approach has been used to develop data driven crowd simulations. The proposed approach in this paper extends this idea to model vehicles in a highway with their unique properties, such as keeping a lane and changing a lane.

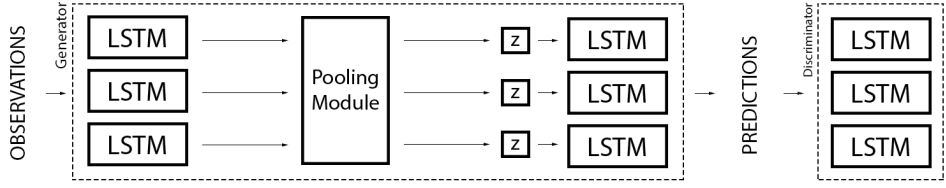

Fig. 1. Architecture of the trajectory generator

TABLE I
non-Deterministic Trajectory Generator Parameters

| | |
|---|---|
| Length of observation sequence, $o_l$ | 8 |
| Length of prediction sequence, $p_l$ | 8 |
| Embedding MLP node size, $s_{MLP}$ | 64 |
| Sampling time, $t_s$ | 0.1s |

The time series GAN formulation can be noted as follows:

$$e_i = \phi(x_i^t, y_i^t) \quad (4)$$
$$h_{ei}^t = Encoder(h_{ei}^{t-1}; e_i^t) \quad (5)$$
$$c_i = \gamma(P_i; h_{ei}^t) \quad (6)$$
$$h_{di}^t = [c_i; z] \quad (7)$$
$$(\hat{x_i^{t+1}}, \hat{y_i^{t+1}}) = Decoder(h_{di}^t) \quad (8)$$

where $x_i^t$ and $y_i^t$ represents the position of the vehicle $i$ at time $t$. $e$ stands for the embedded features, $h_{ei}^t$ are the hidden states of the encoder, $P_i$ is the above mentioned social component and finally $h_{di}^t$ is the hidden states of the decoder. $\hat{x_i^{t+1}}$ and $\hat{y_i^{t+1}}$ are the generated new positions, $\gamma$ and $\phi$ represent a non-linear activation function such as ReLU.

### A. Generating non-Deterministic Trajectories for Vehicles

To introduce randomness factor and simulate faulty driver behaviors, classic vehicle controllers have been replaced with a semi-supervised trajectory generator. Trajectory generator used in this work has been built on Social-GAN [21] architecture, which is a generative adversarial network that generates data based on past observations. Trajectory generator uses recurrent neural network [22] type architecture LSTM to preserve time-dependent relations and a pooling module to aggregate information across subjects. Generator is based on encoder-decoder framework where there is a pooling module between them. The network embeds locations of all subjects with the help of a single layer MLP with the node size $s_{MLP}$. It observes a sequence of length $o_l$ and predicts the next sequence from now of length $p_l$.

Training of the trajectory generator optimizes the network based on its Final Displacement Error ($FDE$) and Average Displacement Error ($ADE$). It has been observed that if $p_l$ selected to be smaller, the network converges faster. Since there are fewer steps to predict on, chance of doing false estimations for the model becomes smaller. However, using few steps also prevents predicting long-term movements of the vehicles. Since the dynamic simulation environment iterates all the variables at each time step, a predictor has been used to optimize next 8 positions, but only the first one is used in each simulation time step. NGSIM [23] vehicle trajectory data have been used to train the non-deterministic trajectory generator network.

Since the data has many vehicles at each time-step, the cars in the data had to be grouped. K-means clustering [24] has been used to group the vehicles by their locational proximities. The groups have been fed separately into the network during the training phase.

As in Figure 2, the trajectory-generator can predict the trajectory types of such as completing a lane-change, starting a lane-change, going steady with only taking 8 samples from NGSIM vehicles.

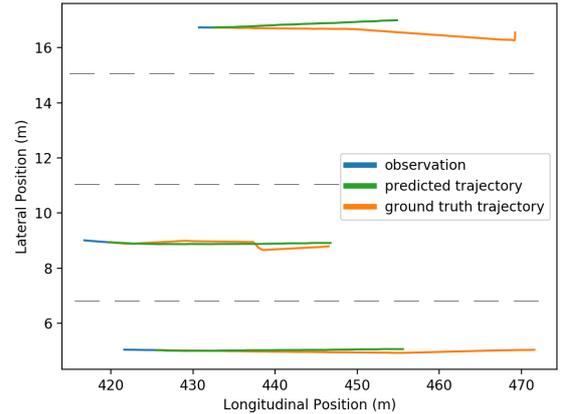

Fig. 2. Performance check of the trajectory generator

Since the network is trained on observation sequences of length 8 (0.8s), it is not possible to extract exact pattern of lane changes and accelerations. These types of actions were added into generated trajectories manually, which makes the trajectory generator network semi-supervised.

A recent work about the lane-changing behaviors of the vehicles [25] gives a formula that directly calculates mean lane-changing frequency of a vehicle from the collected data;

$$SrLC = \frac{n}{q} \times \frac{1000}{L} \quad (9)$$

where $n$ denotes number of observed lane-changes, $q$ denotes the total unique vehicle count, $L$ denotes the length of the observed road in meters. It gives a frequency with unit $\left(veh^{-1}km^{-1}\right)$.

A custom-defined lane-change decision function has been developed based on the extracted mean lane-change characteristics from NGSIM data.

$$p(\text{lane-change}|t) = \frac{t}{t_m} \quad (10)$$

where $t$ denotes current time step and $t_m$ denotes mean frequency of lane-change action in time steps. When a lane-change action finishes, $t$ value will be reset to 0.

Another work about the acceleration and lane-changing dynamics of vehicles in NGSIM [26] gives an approximation method to find a mean lane-changing duration with a standard deviation for given dataset.

$$T_{lc} = \tau_e + \tau_s \quad (11)$$

The mean lane change duration $\overline{T}_{lc}$ is calculated using Eq. (11). $\tau_s$ denotes the relative start time of the lane-change, $\tau_e$ denotes the end time of the lane-change. A Gaussian distribution was fit to lane-change speeds based on calculated mean and standard deviation. This distribution has been used for setting the speed in the phase of lane-change in the simulation.

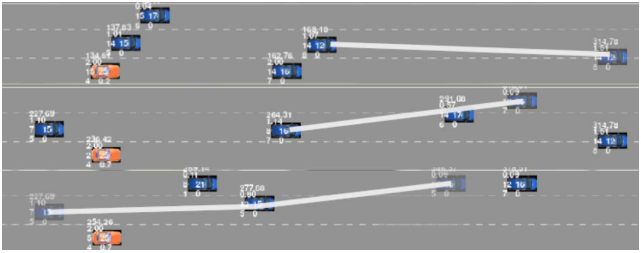

Fig. 3. Generated Trajectories in the Simulation

The Figure 3 shows the simulation environment and different generated vehicle trajectories. The orange vehicle is the ego vehicle. The trajectories of the blue vehicles have been generated by the developed trajectory-generator. White line shows the trajectory path of a certain vehicle. In the figure, trajectories of the three different vehicles have been shown. 4 sample consequential frames have been selected from a random episode in the simulation environment.Then frames were added on top of each other to track the changes of the position of the vehicles.

## IV. SIMULATION SETUP

In this section, initialization phase, observation and action spaces for the agent, two-point steering model, MOBIL and IDM parameters are given. A system with NVIDIA Tesla V100 GPU and 128GB RAM has been used in training.

### A. Initialization Phase

The highway environment for training the RL agent is controlled by several parameters. The highway is initialized with $n$ number of lanes. Next, $m$ agents are placed in the environment, following certain rules. Each of the agents has a dimension of $4.5 \times 2.5$ meters. The initial longitudinal ($x_0$) and lateral ($y_0$) positions of the vehicles were determined, provided that the maximum initial vehicle spread of the vehicles did not exceed the maximum distance $d_{long}$ and not fall below the minimum inter-vehicle distance $d_{\triangle}$. The agent in the middle was chosen as ego-vehicle when agents were sorted according to their longitudinal positions. The agents in front of the ego-vehicle have relatively lower initial speed $v_0$ within the range of $[v_{min}^{front}, v_{max}^{front}]$. The agents behind the ego-vehicle have relatively higher initial speed $v_0$ within the range of $[v_{min}^{rear}, v_{max}^{rear}]$. The same range layout $[v_{min}^{ego}, v_{max}^{ego}]$ also applies for the ego vehicle. Also, desired speeds are defined for each agent included ego-vehicle in the range of $[v_{min}^d, v_{max}^d]$ and $v_{ego}^d$. A distance limit $d_{max}$ has been set to finish each episode. These parameters have been determined by taking reference values from [6]. Table II shows the parameters.

TABLE II
SIMULATION PARAMETERS

| | |
|---|---|
| Minimum inter-vehicle distance, $d_{\triangle}$ | $25\ m$ |
| Maximum initial vehicle spread, $d_{long}$ | $200\ m$ |
| Desired speed for ego vehicle, $v_{ego}^d$ | $25\ m/s$ |
| Episode length, $d_{max}$ | $5000\ m$ |
| Desired speed range for other vehicles, $[v_{min}^d, v_{max}^d]$ | $[18, 26]\ m/s$ |
| Rear vehicles initial speed range, $[v_{min}^{rear}, v_{max}^{rear}]$ | $[15, 25]\ m/s$ |
| Front vehicles initial speed range, $[v_{min}^{front}, v_{max}^{front}]$ | $[10, 12]\ m/s$ |
| Initial speed range for ego vehicle, $[v_{min}^{ego}, v_{max}^{ego}]$ | $[10, 15]\ m/s$ |
| Number of vehicles, $m$ | 9 |
| Number of lanes, $n$ | 3 |

### B. Observation states and action spaces

The ego vehicle has the capability to observe the entire environment. The table III shows the observable states which were described such that, it can adapt to different number of vehicles which besiege the ego vehicle. [6].

TABLE III
OBSERVATION STATES OF THE EGO VEHICLE

| | |
|---|---|
| $s_1$, | Normalized ego vehicle speed $v_{ego}/v_{ego}^d$ |
| $s_2$, | ego vehicle $\begin{cases} 1, & \text{if there is a lane to the leftt} \\ 0, & \text{otherwise} \end{cases}$ |
| $s_3$, | ego vehicle $\begin{cases} 1, & \text{if there is a lane to the right} \\ 0, & \text{otherwise} \end{cases}$ |
| $s_{3i+1}$, | Normalized relative position of vehicle $i$, $\Delta s_i / \Delta s_{max}$ |
| $s_{3i+2}$, | Normalized relative velocity of vehicle $i$, $\Delta v_i / v_{max}$ |
| $s_{3i+3}$, | $\begin{cases} -1, & \text{if vehicle } i \text{ is two lanes to the right of ego vehicle} \\ -0.5, & \text{if vehicle } i \text{ is one lanes to the right of ego vehicle} \\ 0, & \text{if vehicle } i \text{ is in the same lane as the ego vehicle} \\ 0.5, & \text{if vehicle } i \text{ is one lanes to the left of ego vehicle} \\ 1, & \text{if vehicle } i \text{ is two lanes to the left of ego vehicle} \end{cases}$ |

where the maximum allowed speed for all vehicles is $v_{max}$, maximum relative position between ego vehicle and vehicle $i$ is $\Delta s_{max}$ and the maximum allowed speed for ego vehicle is $v_{ego}^d$. The are three action spaces for the vehicle. $a_1$ for no lane change, $a_2$ for left lane change and $a_3$ for right lane change .

*C. Vehicle and Steering Control Model*

To simulate the dynamics of vehicles, non-linear kinematic bicycle model is used. Steering angle $\delta_f$ and the acceleration value $a$ have been set to be control inputs. To calculate $\delta_f$ and $a$, two-point visual control model of steering [27] and the IDM [8] is used, respectively. Steering angle $\delta_f$ with two key-points from the rear and front of the vehicle is estimated by a calculation method called two-point visual control model.

## V. Hyper-parameters

In this work, we have turned our attention on reward function and the neural network architecture which have the greatest effect on the performance of the agent.

*A. Neural network architecture*

2 NoisyLinear layers have been used, which is defined in Rainbow [28] with $\{256\}$ as the number of neurons in 2 hidden layers which are all activated with $ReLU$ activation function. In order to prevent over-fitting and decrease the training time, the architecture has been kept simple. Basic grid search method is used to determine number of neurons in the model.

*B. Driving assistance model hyper-parameters*

TABLE IV
MOBIL Hyper-Parameters

| | |
|---|---|
| Changing threshold, $a_{th}$ | 0.1 m/s2 |
| Politeness factor for rear vehicles, $q$ | 0.5 |
| Politeness factor for side vehicles, $p$ | 1 |
| Maximum safe deceleration, $b_{safe}$ | 4 $m^2$ |

TABLE V
IDM Hyper-Parameters

| | |
|---|---|
| Minimum gap, $d_0$ | 2 $m$ |
| Safe time headway, $T$ | 1.6 $s$ |
| Desired deceleration, $b$ | 1.7 $m/s^2$ |
| Maximum gap for empty lane, $d_{max}$ | 10000 m |
| Minimum deceleration, $a_{min}$ | $-20$ $m/s^2$ |
| Maximum acceleration, $a_{max}$ | 0.7 $m/s^2$ |
| Acceleration exponent, $\delta$ | 4 |

*C. Reward function:*

The objective of this work is to train an agent that can adapt in various environments and drive safely without violating the safety of the road. The parameters used in the reward function are shown bellow.

$$r(s,a,s') = \begin{cases} \text{Speed Reward: } (v_{cur} - 15)/(v_{des} - 15) \\ \text{Low Acc Reward: } - \text{Speed Reward} \\ \text{Lane Change Penalty: } -1 \\ \text{Out of Road Penalty: } -100 \\ \text{Hard Crash: } -100 \\ \text{Soft Crash: } -10 \\ \text{Goal: } +100 \end{cases}$$

There are two different crashes defined in the reward function. Hard crash is the direct collusion with the other vehicle whereas the soft crash is the dangerous approach to the other vehicle.

## VI. Results

After initialization of the simulation environment, two types of RL agents named Agent$_{IDM}$ and Agent$_{GAN}$ have been trained on deterministic traffic scenarios that have been led by IDM-MOBIL algorithms and uncertain traffic scenarios that have been generated by the trajectory generator network, respectively. Agent$_{IDM}$ has been trained for $10,000,000$ iterations. After that, Agent$_{GAN}$ has been trained for $3,000,000$ iterations in the uncertain traffic environment with using transfer learning.

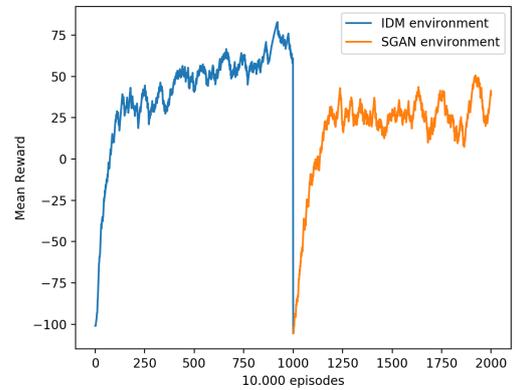

Fig. 4. Comparison of reward in two different training phases

The effect of the environment on the training process is illustrated in Fig. 4. As expected, the agent trained on the static environment shows worse performance at earlier iterations of transfer learning on the environment, which is based on the non-deterministic trajectory generator network. As iterations pass, the model learns to adapt on uncertainties.

TABLE VI
Number of crashes on Traffic$_{IDM}$

| Models | Hard Crash | Soft Crash |
|---|---|---|
| Agent$_{IDM}$ | 19 | 2 |
| Agent$_{GAN}$ | 9 | 0 |

TABLE VII
Performance of Agents on Traffic$_{GAN}$

| Models | Normalized (% MOBIL) | Mean Reward |
|---|---|---|
| Agent$_{IDM}$ | 5.21% | $-22.33 \pm 100.66$ |
| Agent$_{GAN}$ | 114.82% | $33.62 \pm 95.19$ |

The agents have been tested in 2 different types of environments together as shown in Tables VI and VII. The first environment, Traffic$_{IDM}$, is a static environment where the other actors in the environment do not make complex decisions such as changing their lanes. The second environment Traffic$_{GAN}$, is based on the non-deterministic trajectory generator network where other agents acts in a similar way with real traffic scenarios, which can cause them to make unnecessary decisions. Two agents have been compared at the same time with the MOBIL in the Traffic$_{GAN}$ in order to have a fair comparison between two agents. The results in VII have been obtained after 1000 sample simulation runs.

According to Table VI; in a relatively certain and non-complex traffic, even though Agent$_{IDM}$ has been tested in the environment that it has been trained, it stays behind of the Agent$_{GAN}$. Also Agent$_{GAN}$ does relatively good considering it hasn't been trained on the same environment. According to Table VII; In a complex and uncertain traffic, Agent$_{IDM}$ obtains less rewards than the MOBIL algorithm since it hasn't been tested in the environment that it has been trained. Agent$_{GAN}$ does better than MOBIL and Agent$_{IDM}$ since it has observed the uncertain and faulty behavior situations during its training phase.

From the tables VI and VII, It can be claimed that an RL agent that has been trained in a static non-complex environment can not learn the underlying dynamics and can not adapt to uncertainty of the real-world applications where surrounding vehicles make complex or faulty decisions such as lane changing, instant-acceleration or instant-slowing.

Results mentioned above prove that training an RL agent in a complex and uncertain environment yields an agent with better generalization capability.

## VII. CONCLUSIONS

In this work, a deep reinforcement learning agent has been trained to make safe driving decisions in non-deterministic traffic environments which have been developed with a non-deterministic trajectory generator network. We have shown that the trained agent has superior performance in uncertain environments compared with the rule-based methods. The ego vehicle can adapt in different environments and reach its goal without requiring any modification. For the future work, we are planning to add more complex scenarios to the environments in order to make the agent learn different situations such that exiting from a highway in a particular direction.

## VIII. ACKNOWLEDGEMENTS

This work is supported by Istanbul Technical University BAP Grant NO: MOA-2019-42321